\documentclass[10pt,twocolumn,letterpaper]{article}

\usepackage{cvpr}              
\usepackage{adjustbox}
\usepackage{multirow, makecell}
\usepackage{pifont}
\usepackage{algorithm}
\usepackage{algorithmic}
\usepackage{enumitem}
\usepackage{amsmath,amssymb,mathtools}
\usepackage[table]{xcolor}
\usepackage{bm}
\usepackage{url}
\definecolor{cvprblue}{rgb}{0.21,0.49,0.74}
\usepackage[pagebackref,breaklinks,colorlinks,allcolors=cvprblue]{hyperref}

\def\paperID{43828} 
\def\confName{CVPR}
\def\confYear{2026}

\title{RAID: Towards Robust AI-Generated Image Detection with Bit-Reversed Images}

\author{
	Renxi Cheng\textsuperscript{1}, Jie Gui\textsuperscript{1,2,3}, Hongsong Wang\textsuperscript{4,5}\\
    $^{1}$School of Cyber Science and Engineering, Southeast University, Nanjing 210096, China\\
    $^{2}$Purple Mountain Laboratories, Nanjing 210000, China \\
	$^{3}$Engineering Research Center of Blockchain Application, Supervision And Management\\ (Southeast University), Ministry of Education, China \\
	$^{4}$School of Computer Science and Engineering, Southeast University, Nanjing 210096, China \\
	$^{5}$Key Laboratory of New Generation Artificial Intelligence Technology and Its Interdisciplinary \\
	Applications (Southeast University), Ministry of Education, China \\ 
	\tt\small\{renxi, guijie, hongsongwang\}@seu.edu.cn \\
}

\begin{document}
\maketitle

\begin{abstract}
The rapid advancement of image generation models has made it increasingly difficult for people to distinguish AI-generated images from real ones. To prevent the potential risks associated with the misuse of fake images, AI-generated image detection has gained significant attention. Existing methods neglect the inherent differences between real and fake images, thus lacking robustness and generalization ability. In this work, we innovatively investigate AI-generated image detection using bit-planes, and introduce the bit-reversed image. We propose a simple yet effective pipeline consisting of construction of bit-reversed images, gradient-based patch selection and a convolutional classifier. Besides, we provide a theoretical analysis from the mathematical perspective to demonstrate the validity of our approach. We also introduce two challenging datasets for AI-generated image detection. 
Extensive experiments verify the effectiveness of our approach across different settings, including cross-generator generalization, cross-dataset generalization and zero-shot performance. Without bells and whistles, our approach outperforms existing methods on over 40 benchmarks, and is nearly 100 times faster than counterparts. The code is at \url{https://github.com/renxi-seu/RAID}.
\end{abstract}

\section{Introduction}
\label{sec:intro}
The realism of images produced by advanced generative models, such as Generative Adversarial Networks (GANs) \cite{goodfellow2014generative} and Diffusion Models \cite{rombach2022high}, has improved dramatically in recent years. This progress raises serious concerns about the potential misuse of AI-generated images \cite{juefei2022countering}, such as the creation of deceptive or harmful content. 
Such risks underscore the pressing need for robust methods that can accurately differentiate AI-generated images from real ones.

Existing deepfake detection works aim at generalizability~\cite{11494439,11083677}, which can be categorized into three groups: spatial domain-based, frequency domain-based, and patch-based approaches. The first category analyzes pixel-level texture patterns and gradient artifacts~\cite{tan2023learning} and reconstruction error~\cite{wang2023dire}. 
The second reveals artifacts often imperceptible in pixel space by focusing on frequency artifacts~\cite{zhang2019detecting} and high-frequency features~\cite{dzanic2020fourier}. The third approach learns and aggregates features from local patches rather than processing the entire image~\cite{chen2024single,zheng2024breaking,yang2025all}. 
However, these methods employ sophisticated models to learn effective features from images, thereby neglecting to discover the inherent differences between real and fake images.

\begin{figure*}[t]
  \centering
  \small
  \includegraphics[width=1\linewidth]{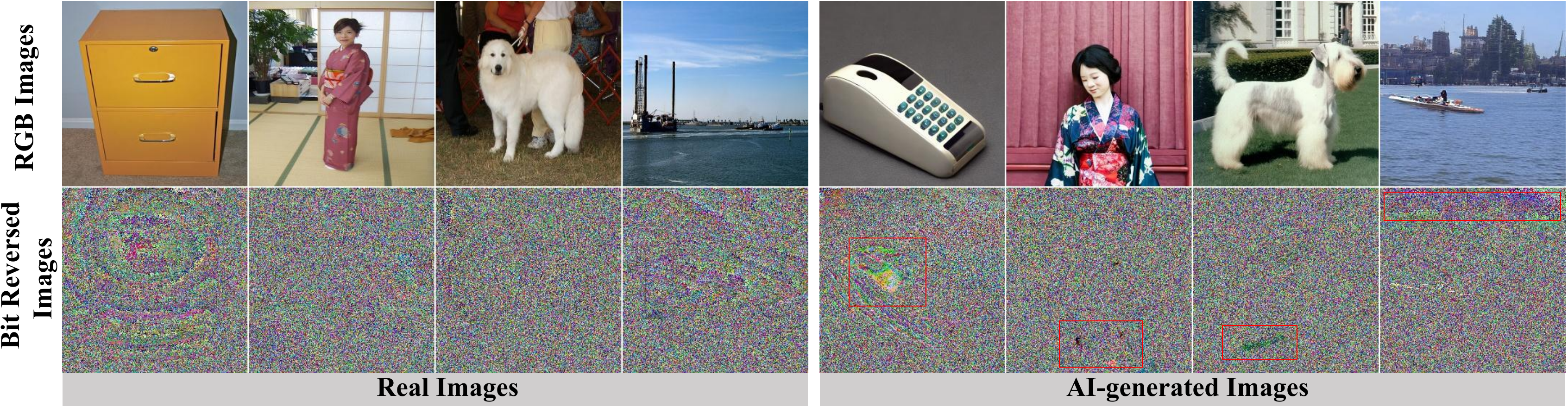}
  \caption{\textbf{Comparison of bit-reversed images between real and fake images.} Noticeable artifacts appear in certain regions of reversed images in fake images, while the noise in bit-reversed images of real images tends to be more naturally distributed.}
  \label{fig:motivation}
\end{figure*}

A grayscale image can be reversibly decomposed into eight bit-planes. Least Significant Bit (LSB) substitution is a well-known technique in the fields of information hiding~\cite{chan2004hiding} and steganography~\cite{elharrouss2020image}. Bit-plane has shown potential and advantages in image stabilization~\cite{ko1998digital}, image encryption~\cite{gan2019chaotic}, image compression~\cite{zhang2024learned}, and implicit neural representations~\cite{han2025towards}. Such bit-plane-based methods are primarily designed for image processing. 
Since bit-planes inherently possess the ability to convey fine details within an image, they hold the potential to discern subtle differences between real and AI-generated images. However, marrying bit-planes and AI-generated image detection, which is a promising direction, has not been studied yet.
In prior research on frequency-based method~\cite{zhong2023patchcraft,tan2024frequency}, operations such as generator upsampling and regularization losses introduce noticeable discrepancies (artifacts) in the frequency domain between generated and real images. These artifacts manifest as unnatural directional properties in the spectrum of synthetic images, with diffusion-generated images often exhibiting energy concentrated in specific frequency bands that deviate from the uniform attenuation model characteristic of natural images. The significant discrepancies between generated and real images across bit-planes particularly in low-order planes provide a strong basis for bit-plane-based Deepfake detection.

To enhance these discrepancies, we innovatively introduce the bit-reversed image, which simply reverses the permutation order of bit-planes before constructing the image. The bit-reversed image has two distinct characteristics. First, it can be reversibly converted into bit-plane images or the original image. Second, the visual content of the original image is encrypted, while the noise and fine details are amplified. Fig.~\ref{fig:motivation} shows the comparison of corresponding bit-reversed images between real and AI-generated images. It can be seen that, for fake images, artifacts are apparent in the corresponding bit-reversed images. 

To this end, we propose a simple yet effective approach for AI-generated image detection. We first synthesize the bit-reversed image from the original image based on bit-planes, and then design a patch-based classifier to detect fake content. Specifically, we investigate both bit-forward image and bit-reversed image during the bit-plane-based image construction. The patch-based classifier consists of the gradient-based patch selection followed by a convolutional classifier. Both construction of bit-reversed images and gradient-based patch selection operate at millisecond-level speed and involve no trainable parameters. 
The convolutional network is adapted to prevent premature feature compression and accommodate small image patches as input.
We assess our approach across various AI-generated image detection settings, including cross-generator evaluation, zero-shot generalization, and cross-dataset evaluation. 
Our approach achieves state-of-the-art performance on more than 40 benchmarks, significantly surpassing existing approaches. In summary, our main contributions are as follows:
\begin{itemize}
\item[$\bullet$] \textbf{Innovative deepfake representation:} We innovatively tackle AI-generated image detection based on bit-planes, and introduce the bit-reversed image that can be reversibly constructed from the original image.
\item[$\bullet$] \textbf{Efficient pipeline design:} We propose a simple yet effective pipeline for AI-generated image detection, which significantly outperforms existing approaches on standard benchmarks while operating at the millisecond level.
\item[$\bullet$] \textbf{New and challenging benchmarks:} Since our approach has nearly saturated existing benchmarks, we introduce two challenging datasets to promote future research.
\end{itemize}

\section{Related Work}
\label{sec:rela}
\textbf{AI-Generated Image Detection:} 
Existing methods can be roughly categorized into spatial domain-based, frequency domain-based, patch-based, and multimodal-based approaches~\cite{11456362}. In contrast, this work is among the first bit-plane-based approaches.

For spatial domain-based methods, GLFF~\cite{ju2023glff} utilizes a fusion of global and local features to effectively capture inconsistencies at multiple scales. DIRE~\cite{wang2023dire} leverages reconstruction error as a fundamental detector for AI-generated images. LaRE$^2$~\cite{luo2024lare} incorporates refinement mechanisms for both spatial and channel features to boost feature learning. CoD~\cite{jia2025secret} exploit color distribution inconsistencies via quantization–restoration analysis.

For frequency domain-based methods, Corvi et al.~\cite{corvi2023detection, corvi2023intriguing} extend spectral analysis to diffusions by identifying unique frequency fingerprints. PatchCraft~\cite{zhong2023patchcraft} postulates that artifacts predominantly manifest in high-frequency texture regions. FreqNet~\cite{tan2024frequency} compels the model to learn source-agnostic features by incorporating high-frequency representation modules and frequency convolution layers into the CNN classifier.
SPAI~\cite{karageorgiou2025any} introduce a spectral learning-based method based on frequency reconstruction and reconstruction similarity.

\begin{figure*}[t]
  \centering
  \small
  \includegraphics[width=0.92\linewidth]{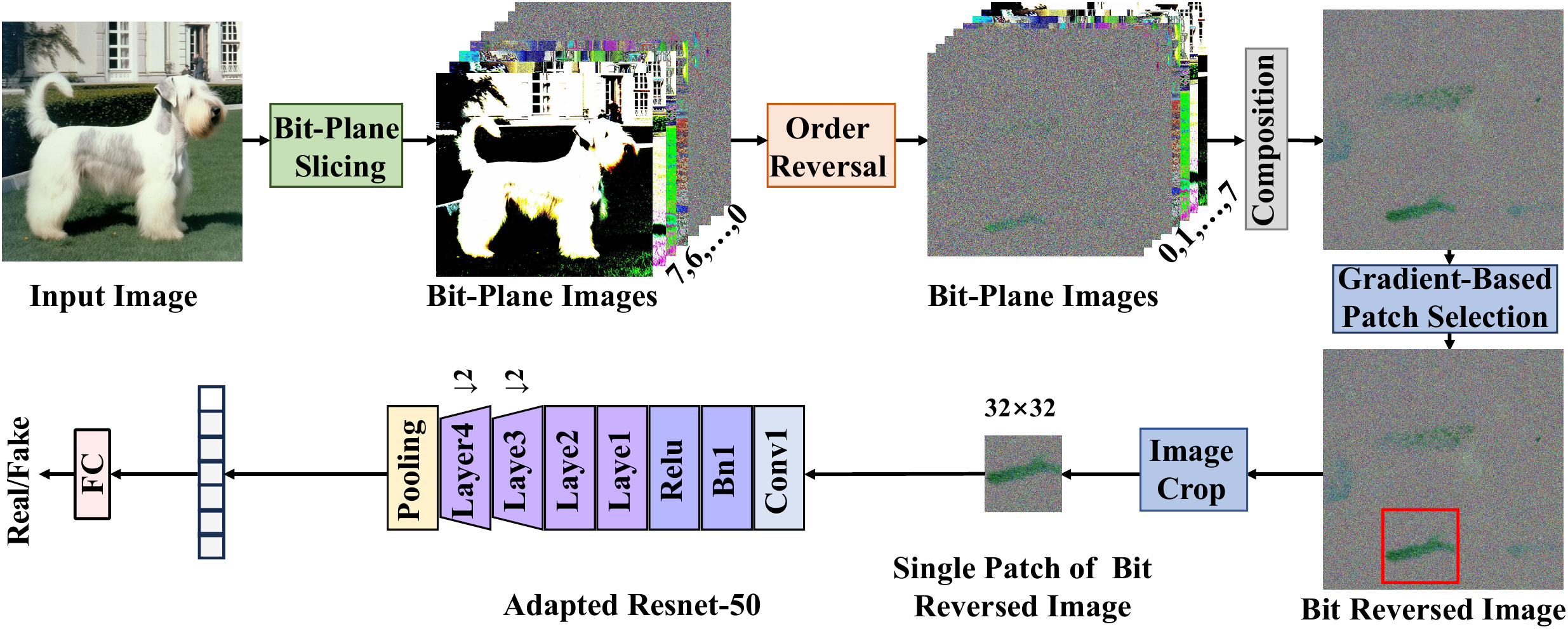}
  \caption{\textbf{Pipeline for the proposed method. } Given an RGB image, our approach first extracts the bit-reversed image, then heuristically selects a patch, and finally performs fake classification using the adapted ResNet-50 (↓2 indicates a 2× downsampling of spatial resolution in feature learning).}
  \label{fig:method}
\end{figure*}

As for patch-based methods, PatchCraft~\cite{zhong2023patchcraft} regards the difference between the patches with the highest and lowest diversity as the detection criterion, while SSP~\cite{chen2024single} selects the patch with the highest diversity to expose artifacts. Other studies challenge the adequacy and effectiveness of relying solely on a single patch or several patches. Zheng et al.~\cite{zheng2024breaking} present a classifier trained on patch-shuffled images and aggregates patch-wise features. Yang et al.~\cite{yang2025all} randomly replace partial patches with real patches to force the model to learn artifacts from all patches. Xiao et al.~\cite{xiaohigh} show that high-quality AI-generated image detection can be improved by selecting patches identified through low-level visual cues. 

\noindent\textbf{Bit-Plane-Based Image Processing:} 
Bit-plane-based operations play a significant role in image processing. A notable example is Least Significant Bit (LSB) substitution~\cite{chan2004hiding}, which is a simple and widely adopted data hiding method. Bit-plane-based methods can be used for reversible data hiding in encrypted images by exploiting intra- and inter-bit-plane correlations or using asymmetric coding~\cite{kumar2023bit, zhang2024reversible}. 
They can also be applied to image stabilization~\cite{ko1998digital} and image encryption~\cite{gan2019chaotic}. Punnappurath et al.~\cite{punnappurath2021little} propose a bit-plane-wise deep learning framework for bit-depth reconstruction that progressively recovers residuals at each bit-plane level. Zhang et al.~\cite{zhang2024learned} present bit plane slicing with a dimension-tailored autoregressive model to enhance latent variable use and improve lossless image compression efficiency. Han et al.~\cite{han2025towards} introduce a bit-plane decomposition approach for implicit neural representations, enabling faster convergence and lossless fitting of high bit-depth signals. LOTA~\cite{wang2025lota} introduces a bit-planes guided noisy image generation for AI-generated image detection.
Different from these works, and similar to LOTA, we study a bit-plane–based method for detecting AI-generated images.

\section{Methodology}
\label{sec:method}
We address AI-generated image detection from the perspective of bit-planes. The pipeline of our approach RAID is shown in Fig.~\ref{fig:method}. For each module of RAID, theoretical proofs from the mathematical perspective are also provided. Details are described below. 

\begin{figure*}[t]
  \centering
  \small
  \includegraphics[width=1\linewidth]{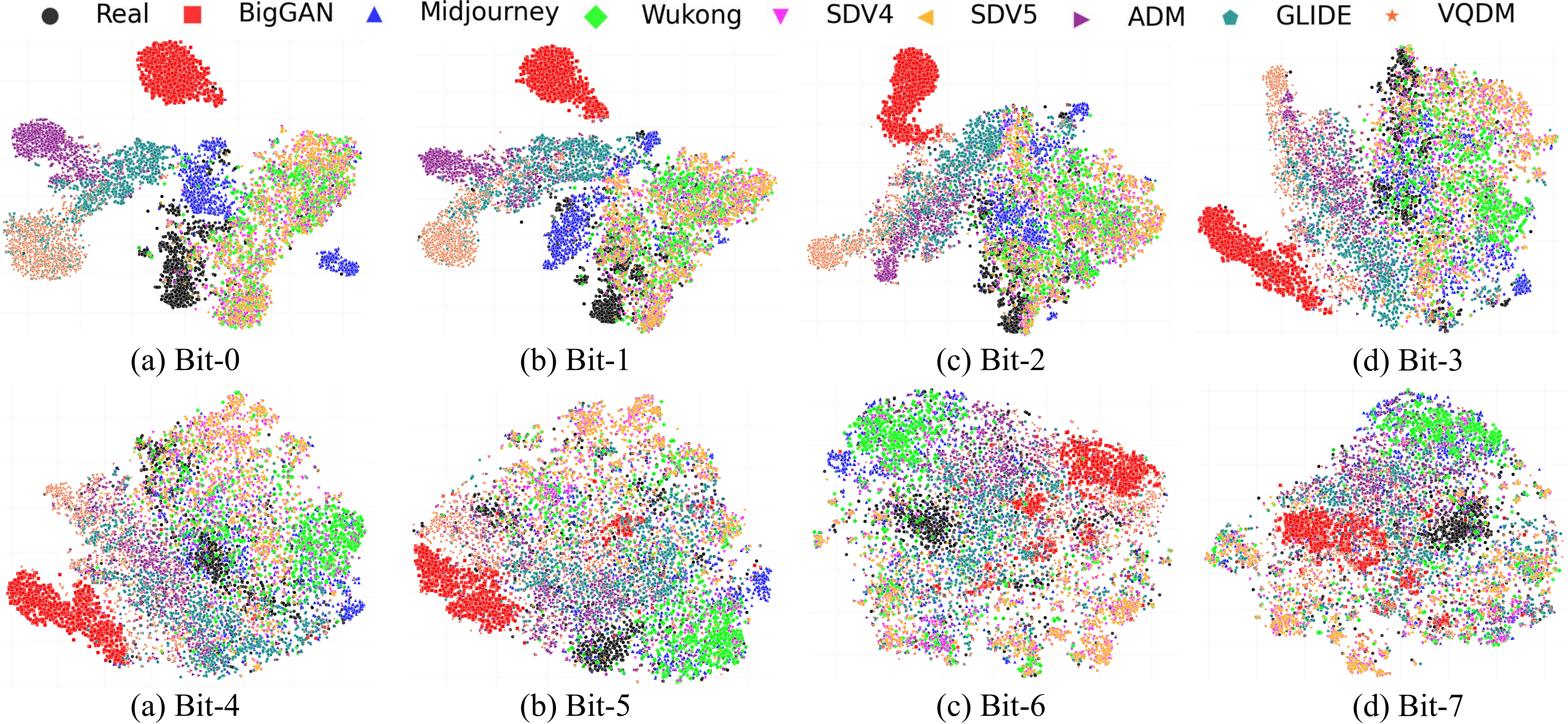}
  \caption{\textbf{Comparison of sample distributions between real and fake images of different bit-planes.} For low bit-planes, especially bit-0, bit-1, and bit-2, the distributions of real and fake images are clearly separable, and different types of generators also exhibit distinct sample distributions. In contrast, for high bit-planes, the distributions of real and fake images heavily overlap.}
  \label{fig:pca_bit_feat}
\end{figure*}

\subsection{Construction of Bit-Reversed Images}
A grayscale image $\bm{I}$ can be losslessly decomposed into eight binary images using bit-plane decomposition, each representing a specific bit-plane. For the RGB image, each channel $c=\{R,G,B\}$ corresponds to a gray-scale image $I$, which can be subsequently decomposed as 8 bit-planes $\{\bm{x_k^c}|k=0,1,...,7\}$. As the value of $k$ increases, higher bit-planes capture the semantic structural information, similar to low-frequency components. As the value of $k$ decreases, lower bit-planes contain noise patterns, analogous to high-frequency components.

Although AI-generated images from current generators look almost indistinguishable from real images, significant discrepancies exist between AI-generated and real images in the low bit-planes, as illustrated in Fig.~\ref{fig:pca_bit_feat}. As AI-generated images lack true physical sensor noise, they show structured or unnatural randomness in low-bit planes. Since the neural networks learn semantic correlations during image synthesis, semantic structures may leak into low bit-planes of AI-generated images. This evidence can be observed in Fig.~\ref{fig:motivation} and is also confirmed in LOTA~\cite{wang2025lota}.

To highlight invisible artifacts in low bit-planes of the AI-generated image, a straightforward idea is to reverse the order of eight bit-planes before recomposing the image. The composed image is computed as:
\begin{equation}
  \tilde{\bm{I}} = \sum\limits_{k = 0}^7 {w_k} \cdot \bm{x_k^{c}},
  \label{eq:decomposition}
\end{equation}
where $w_k$ denotes the weight of the $k$-th bit-plane. The eight weights can be written as a weight vector $\bm{w} = [\bm{w_0}, \bm{w_1}, \bm{w_2}, \bm{w_3}, \bm{w_4}, \bm{w_5}, \bm{w_6}, \bm{w_7}]$.

\begin{figure*}[t]
  \centering
  \small
  \includegraphics[width=1\linewidth]{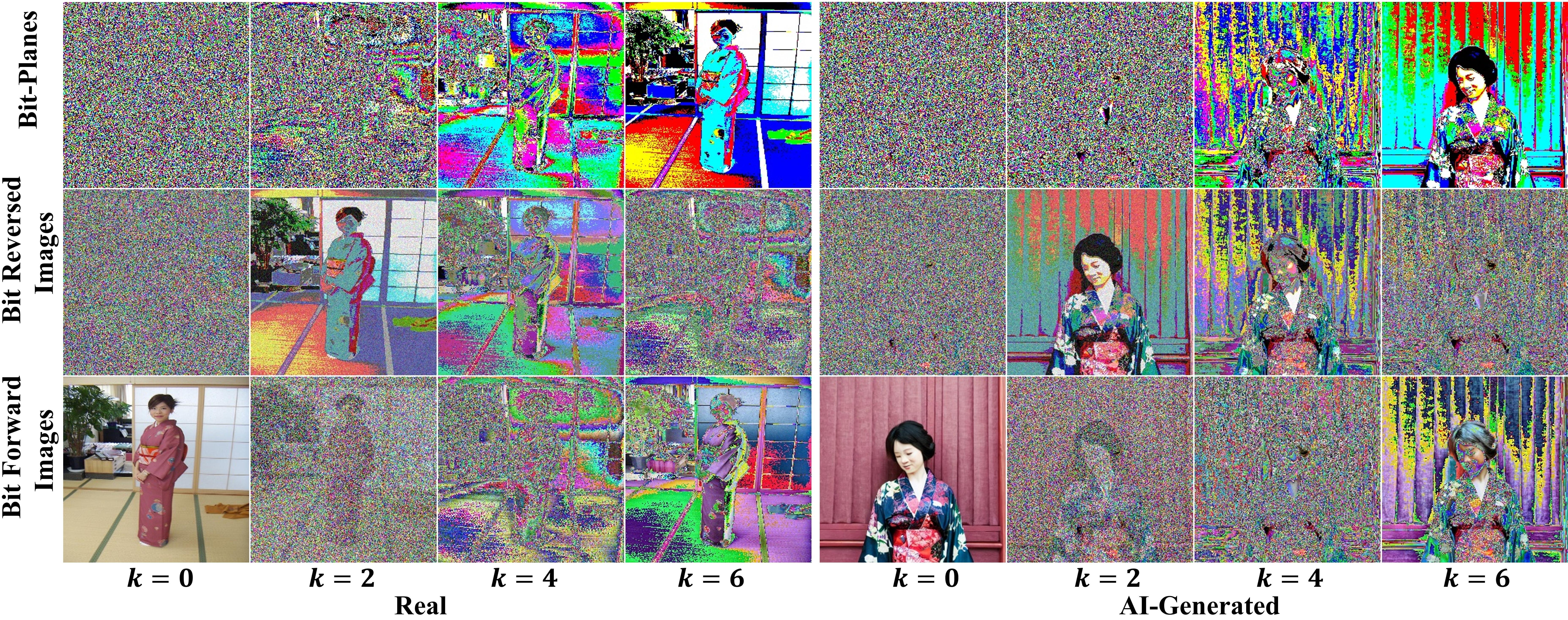}
  \caption{\textbf{Visualizations of bit-planes, bit-reversed and bit-forward images for real and fake RGB images}. $k = \left\{ 0, 1, \ldots, 7 \right\}$ is index of bit-planes. For bit-forward or bit-reversed images, $k$ denotes the positions of the left circular shift operation. For AI-generated images, different values of $k$ result in varying artifacts.}
  \label{fig:imethod2}
\end{figure*}

For constructed images, varying the value of $\bm{w}$ yields composed images of different styles. These images can be categorized into two types: Bit-Forward Images and Bit-Reversed Images, as visualized in Fig.~\ref{fig:imethod2}.

\noindent\textbf{Bit-Forward Images:} The default weight vector is: $\bm{w}=[2^0,2^1, 2^2, 2^3, 2^4, 2^5, 2^6, $
$2^7]$. The weights are determined by the positions of the bit-planes. Specifically, for the $k$-th bit-plane, the corresponding weight is $2^k$. The original image can be equivalently recovered using these weights. 

Given a weight vector with eight elements, performing a left circular shift by one position iteratively yields eight distinct vectors. The transformed weight vector after a left circular shift by $k$ positions is: $\bm{w}=[2^k,2^{k+1}, \dots, 2^7, 2^0,\dots,2^{k-1}]$.

\noindent\textbf{Bit-Reversed Images:} The bit-reversed image (BRI) can be obtained by reversing the order bit-planes, and the corresponding weight vector is: $\bm{w}=[2^7, 2^6, $
$2^5, 2^4, 2^3, 2^2, 2^1, 2^0]$. For the $k$-th bit-plane, the corresponding weight is $2^{7-k}$. Since higher bit-planes have smaller weights, the bit-reversed image can amplify fine details and noise.

Similarly, a left circular shift can be applied to generate eight different weight vectors. The resulting vector after a left circular shift by $k$ positions is: $\bm{w}=[2^{7-k}, 2^{6-k}, \dots, 2^0, 2^7, \dots, 2^{8-k}]$.

\subsection{Patch-Based Classifier}

After constructing bit-reversed images, we design a patch-based classifier that consists of Gradient-Based Patch Selection and a Convolutional Classifier.

\noindent\textbf{Gradient-Based Patch Selection:} Although artifacts in bit-reversed images serve as a critical feature for distinguishing real and generated images, they still contain a lot of irrelevant information that may interfere with detection. To mitigate such interference and amplify the artifacts, we introduce Gradient-Based Patch Selection (GBPS) to select the most informative patch. 

Given the bit-reversed image $\tilde{\bm{x}}^{c}$, we partition it randomly into non-overlapping patches. To evaluate the sparsity of image gradients along various directions, we propose a divergence-based scoring function. For a noisy patch $\tilde{z}_p$, where $p$ represents the patch index, the score $g_p$ is calculated as follows:
\begin{equation}
g_p = \sum_{d \in \mathcal{D}} \left\| \tilde{\bm{x^{c}}} * \bm{g_d} \right\|_1,
\end{equation}
where $*$ denotes the image convolution operation, $\|\cdot\|_1$ represents the $L1$ norm of the matrix and $\mathcal{D} = \{x, y, xy, yx\}$ defines the set of gradient directions. The convolution kernels $\bm{g_x}, \bm{g_y}, \bm{g_{xy}}$ and $\bm{g_{yx}}$ are described as:
\begin{equation}
\begin{aligned}
\bm{g_x} &= \begin{bmatrix} -1 & 1 \end{bmatrix}, \quad\,
\bm{g_y} &= \bm{g_x}^T, \\
\bm{g_{xy}} &= \begin{bmatrix} -1 & 0 \\ 0 & 1 \end{bmatrix}, \quad
\bm{g_{yx}} &= \begin{bmatrix} 0 & -1 \\ 1 & 0 \end{bmatrix}.
\end{aligned}
\end{equation}

The score measures gradients in horizontal, vertical, and diagonal directions. High scores typically correspond to regions with strong high-frequency variations, which are more likely caused by noise or structural details rather than meaningful image content. In AI-generated images, such high-divergence areas often indicate artifacts resulting from imperfections in the generative models. Therefore, we select the noisy patch with the highest $g_p$ score:
\begin{equation}
\bm{\tilde{z}_{p^*}} = \arg\max_{p} g_p,
\end{equation}
where $p^*$ denotes the index of the best image patch.

Although PatchCraft~\cite{zhong2023patchcraft}, ESSP~\cite{chen2024single} and our GBPS all involve patch-based selection, ours distinguishes them in three main aspects. First, GBPS relies on a gradient-based score instead of evaluating texture diversity. Second, our formulation is efficiently implemented through image convolution. Finally, our approach identifies the patch with the maximum score, whereas ESSP opts for the minimum and PatchCraft employs more than a single patch.

\noindent\textbf{Patch-Based Convolutional Classifier:} After selecting the important image patch, we feed it into a ResNet-50–based convolutional classifier~\cite{he2016deepresnet}, chosen for its simplicity and effectiveness. To adapt ResNet-50 for 32$\times$32 patches, we introduce several modifications to preserve spatial information and prevent premature feature compression. Specifically, we reduce the stride of the initial convolution from 2 to 1, keeping the output resolution at 32$\times$32, and remove the max-pooling layer to avoid downsampling to 16$\times$16. We further modify the first bottleneck block in the second layer by changing the strides of both the 3$\times$3 and the corresponding 1$\times$1 convolutions from 2 to 1, ensuring the output remains 32$\times$32. The third and fourth layers remain unchanged, so the adapted ResNet-50 produces 8$\times$8 features before the final pooling layer. By alleviating aggressive early downsampling, these modifications preserve fine spatial details essential for accurate representation learning while maintaining the hierarchical feature extraction capacity of ResNet-50.

\subsection{Theoretical Analysis}

To demonstrate the validity of our approach, we provide a theoretical analysis from the mathematical perspective. We prove Propositions 1 and 2 to demonstrate the effectiveness of BRI, and establish Proposition 3 to justify GBPS.

We model an image $\bm{I}$ as the sum of three independent components: semantic contents $\bm{S}$, forensic micro-structures or artifacts $\bm{A}$ and physical random noise $\bm{R}$, expressed as:
\begin{equation}
  \bm{I}(x, y) = \bm{S}(x, y) + \bm{A}(x, y) + \bm{R}(x, y),
  \label{eq:component}
\end{equation}
where $(x, y)$ is the 2-D pixel coordinates. For real images, the value of the forensic micro-structure $\bm{A}(x,y)$ is approximately zero.

\noindent\textbf{Proposition 1:} The semantic content $\bm{\tilde{S}(x,y)}$ is significantly weakened in the bit-reversed image $\tilde{\bm{I}}(x,y)$.

\noindent\textbf{Proof:} According to the principle of image bit-plane decomposition, the $\bm{I}$, $\bm{S}$ and $\bm{R}$ of a bit-forward image can all be decomposed as follows:

\begin{equation}
  \bm{I}(x,y) = \sum\limits_{k = 0}^7 {2^{k}} \cdot \bm{s_k^{c}} + \sum\limits_{k = 0}^7 {2^{k}} \cdot \bm{a_k^{c}} + \sum\limits_{k = 0}^7 {2^{k}} \cdot \bm{r_k^{c}},
  \label{eq:prove1}
\end{equation}

\noindent where $\bm{s_k^{c}}$, $\bm{a_k^{c}}$ and $\bm{r_k^{c}}$ represent the $k$-th bit-plane in the R channel of $\bm{S}(x,y)$, $\bm{A}(x,y)$ and $\bm{R}(x,y)$.

After reversing the order bit-planes, the $\bm{I}$, $\bm{S}$ and $\bm{R}$ of a bit-reversed image can all be decomposed as follows:

\begin{equation}
  \tilde{\bm{I}}(x,y) = \sum\limits_{k = 0}^7 {2^{7-k}} \cdot \bm{s_k^{c}} + \sum\limits_{k = 0}^7 {2^{7-k}} \cdot \bm{a_k^{c}} + \sum\limits_{k = 0}^7 {2^{7-k}} \cdot \bm{r_k^{c}}.
  \label{eq:prove2}
\end{equation}

Compare Eqs.~(\ref{eq:prove1}) and~(\ref{eq:prove2}), we find that as $k$ increases, $2^k$ and $\bm{s_k^{c}}$ increase monotonically, while $2^{7-k}$ decreases monotonically. Consequently, 
\begin{equation}
  \sum\limits_{k = 0}^7 {2^{7-k}} \cdot \bm{s_k^{c}} < \sum\limits_{k = 0}^7 {2^{k}} \cdot \bm{s_k^{c}}.
  \label{eq:prove3}
\end{equation}
Therefore, the semantic content in bit-reversed images $\tilde{\bm{I}}(x,y)$ is significantly weakened compared to original images.

\noindent\textbf{Proposition 2:} For the bit-reversed image $\tilde{\bm{I}}(x,y)$, $\tilde{\bm{A}}(x,y)\neq0$.

\noindent\textbf{Proof:} During the generation of a fake image, due to the influence of processes such as upsampling and denoising, we have:
\begin{equation}
\bm{I}(x,y)=\bm{S}(x,y)+\delta,
\end{equation}
where $\delta\neq0$. Since $\bm{R}(x,y)=0$ for generated images, it follows that $\bm{A}(x,y)\neq0$ (\textit{i.e.}, $a_k^c\neq0$). Thus, $\tilde{\bm{A}}(x,y)\neq0$, which means the micro-structure of fake images is inevitably present.

\noindent\textbf{Proposition 3:} For the bit-reversed image $\tilde{\bm{I}}(x,y)$, the sum of forensic micro-structure in certain region achieve the maximum value if and only if the gradient of the patch $\nabla\tilde{\bm{I}}(x, y)$ reaches its maximum.

\noindent\textbf{Proof:} Taking the gradient of Eq.~(\ref{eq:prove2}), we have:
\begin{equation}
  \nabla\tilde{\bm{I}}(x, y) = \nabla\tilde{\bm{S}}(x, y) + \nabla\tilde{\bm{A}}(x, y) + \nabla\tilde{\bm{R}}(x, y).
  \label{eq:prove3.1}
\end{equation}
Since $\tilde{\bm{S}}(x, y)$ is very small after bit-reversion and $\tilde{\bm{R}}(x, y)=0$, the gradients $\nabla\tilde{\bm{S}}(x, y)$ and $\nabla\tilde{\bm{R}}(x, y)$ are negligible. Therefore, $\nabla\tilde{\bm{I}}(x, y)$ is primarily dominated by $\nabla\tilde{\bm{A}}(x, y)$, yielding:
\begin{equation}
  \sum\limits_{(x,y)\in S}\nabla\tilde{\bm{I}}(x, y) \approx \sum\limits_{(x,y)\in S}\nabla\tilde{\bm{A}}(x, y),
  \label{eq:prove3.2}
\end{equation}
where $S$ denotes region of the selected patch.
Furthermore:
\begin{align}
  \sum\limits_{(x,y)\in S}\tilde{\bm{A}}(x, y) = \sum\limits_{(x,y)\in S}\nabla\tilde{\bm{A}}(x, y)\text{d}s \nonumber \\
  = S\cdot\sum\limits_{(x,y)\in S}\nabla\tilde{\bm{A}}(x, y) \nonumber = S\cdot\sum\limits_{(x,y)\in S}\nabla\tilde{\bm{I}}(x, y),
  \label{eq:prove3.3}
\end{align}
where $S$ is a constant. It is evident that the sum of forensic micro-structure in certain region achieve the maximum value if and only if the gradient of the patch $\nabla\tilde{\bm{I}}(x, y)$ reaches its maximum.

\begin{table}[htbp]
  \centering
    \resizebox{1\linewidth}{!}{
    \begin{tabular}{lc|ccc|c}
    \toprule
    \multicolumn{2}{c}{Dataset} & Subset & Generator & Label & Images \\
    \hline
    \multirow{10}{*}{GID} & \multirow{2}{*}{Imagen 2}
    & Imagen 2 & Google Imagen 2 & Fake & 2,000 \\
    & & Real & — & Real & 2,000 \\
    \cmidrule(lr){2-6}
    & \multirow{2}{*}{FLUX.1}
    & FLUX.1 & FLUX.1 & Fake & 2,000 \\
    & & Real & — & Real & 2,000 \\
    \cmidrule(lr){2-6}
    & \multirow{2}{*}{DALL-E 3}
    & DALL-E 3 & DALL-E 3 & Fake & 2,000 \\
    & & Real & — & Real & 2,000 \\
    \cmidrule(lr){2-6}
    & \multirow{2}{*}{SD3}
    & SD3 & Stable Diffusion 3 & Fake & 2,000 \\
    & & Real & — & Real & 2,000 \\
    \cmidrule(lr){2-6}
    & \multirow{2}{*}{WANX 2.1}
    & WANX 2.1 & WANX 2.1 & Fake & 2,000 \\
    & & Real & — & Real & 2,000 \\
    \hline
    \multirow{10}{*}{GVD} & \multirow{5}{*}{Group 1}
    & MuseV & MuseV & Fake & 10,000 \\
    & & SVD & Diffusion & Fake & 10,000 \\
    & & CogV & CogVideo & Fake & 10,000  \\
    & & Mora & Mora & Fake & 10,000 \\
    & & HD-VG & — & Real & 40,000 \\
    \cmidrule(lr){2-6}
    & \multirow{5}{*}{Group 2}
    & COG & CogVideo & Fake & 2,500  \\
    & & T2VZ & \footnotesize Text2Video-Zero & Fake & 2,500 \\
    & & TAV & \footnotesize Tune-A-Video & Fake & 2,500 \\
    & & VC & \footnotesize VideoCrafter & Fake & 2,500  \\
    & & YT-BI & — & Real & 10,000 \\
    \bottomrule
    \end{tabular}
    }
    \caption{\textbf{A summary of the introduced datasets: GID and GVD.} GID comprises five subsets: Google Imagen 2 (Imagen 2), FLUX.1, DALL-E 3, Stable Diffusion 3 (SD3), and WANX 2.1. GVD comprises two groups, and each group contains four fake subsets and one real subset.} 
    \label{tab:challenging}
\end{table}

\begin{table*}[htbp]
    \centering
    \resizebox{\textwidth}{!}{
    \renewcommand{\arraystretch}{0.8}
    \scriptsize
    \begin{tabular}{lcccccccccc}
        \toprule
        \multirow{2}{*}{Method} & \multirow{2}{*}{AIGCDB} & \multicolumn{9}{c}{GenImage}  \\ 
        \cmidrule(lr){3-11} 
          & & BigG & Midj & Wuk & SDV4 & SDV5 & ADM & GLI & VQDM & Avg.\\
        \midrule
        Spec~\cite{zhang2019detecting} &\cellcolor{gray!35}- &49.8  &52.0  &94.8  &99.4  &99.2  &49.7  &49.8  &55.6  &\cellcolor{gray!35}68.8 \\
        DeiT-S~\cite{touvron2021training} &\cellcolor{gray!35}- &53.5  &55.6  &98.9  &99.0  &99.8  &49.8  &58.1  &56.9  &\cellcolor{gray!35}71.6 \\
        LGrad~\cite{tan2023learning} &\cellcolor{gray!35}75.3 &- &- &- &- &- &- &- &- &\cellcolor{gray!35}- \\
        LNP~\cite{liu2022detecting} &\cellcolor{gray!35}83.8  &- &- &- &- &- &- &- &- &\cellcolor{gray!35}-  \\
        CNNSpot~\cite{wang2020cnn} &\cellcolor{gray!35}70.8  &46.8  &52.8  &78.6  &96.3  &95.9  &50.1  &39.8  &53.4 &\cellcolor{gray!35}64.2  \\
        GramNet~\cite{liu2020global} &\cellcolor{gray!35}68.4 &51.7  &54.2  &98.9  &99.2  &99.1  &50.3  &54.6  &50.8  &\cellcolor{gray!35}69.9   \\
        ResNet-50~\cite{he2016deepresnet} &\cellcolor{gray!35}69.7 &52.0  &54.9  &98.2  &\textbf{99.9}  &99.7  &53.5  &61.9  &56.2  &\cellcolor{gray!35}72.1  \\
        GenDet~\cite{liu2020global} &\cellcolor{gray!35}- &75.0  &\underline{89.6}  &92.8  &96.1  &96.1  &58.0  &78.4  &66.5  &\cellcolor{gray!35}81.6  \\
        Swin-T~\cite{liu2021swin} &\cellcolor{gray!35}- &57.6  &62.1  &\underline{99.1}  &\textbf{99.9}  &99.8  &49.8  &67.6  &62.3  &\cellcolor{gray!35}74.8   \\
        F3Net~\cite{qian2020thinkingF3Net} &\cellcolor{gray!35}- &49.9  &50.1  &\textbf{99.9}  &\textbf{99.9}  &\textbf{99.9}  &49.9  &50.0  &49.9  &\cellcolor{gray!35}68.7 \\
        UnivFD~\cite{ojha2023towards} &\cellcolor{gray!35}78.4  &80.3  &73.2  &75.6  &84.2  &84.0  &55.2  &76.9  &56.9  &\cellcolor{gray!35}73.3  \\
        PatchCraft~\cite{zhong2023patchcraft} &\cellcolor{gray!35}89.3  &72.4  &79.0  &89.3  &89.5  &89.3  &77.3  &78.4  &83.7  &\cellcolor{gray!35}82.3 \\
        DIRE~\cite{wang2023dire} &\cellcolor{gray!35}67.9 &72.6 &58.5 &58.5 &99.2 &95.4 &61.6 &79.3 &49.8 &\cellcolor{gray!35}72.1  \\
        LaRE$^2$~\cite{luo2024lare} &\cellcolor{gray!35}54.2 &63.4 &84.9 &83.7 &99.1 &99.0 &90.8 &\underline{92.0} &64.0 &\cellcolor{gray!35}84.5 \\
        ESSP~\cite{chen2024single} &\cellcolor{gray!35}50.1 &73.9 &82.6 &98.6 &99.2 &99.1 &\underline{78.9} &88.9 &\underline{96.0} &\cellcolor{gray!35}\underline{89.7}  \\
        AIDE~\cite{yan2024sanity} &\cellcolor{gray!35}92.8  &66.9 &79.4 &98.7 &\underline{99.7} &\underline{99.8} &78.6 &91.8 &80.3 &\cellcolor{gray!35}86.9  \\
        VIB-Net~\cite{zhang2025towards}  &\cellcolor{gray!35}- &\underline{95.8} &61.3 &75.9 &71.6 &70.0 &71.5 &69.4 &86.7 &\cellcolor{gray!35}84.2 \\
        UniFD~\cite{ojha2023towards} &\cellcolor{gray!35}77.1 &90.0 &56.1 &70.7 &63.6 &63.9 &67.6 &62.7 &85.6 &\cellcolor{gray!35}70.1 \\
        NPR~\cite{tan2024rethinking}  &\cellcolor{gray!35}91.7 &80.7 &91.7 &94.0 &94.4 &94.4 &87.8 &93.2 &88.7 &\cellcolor{gray!35}90.6 \\
        C2P~\cite{tan2025c2p}  &\cellcolor{gray!35}\textbf{96.2} &98.7 &88.2 &98.8 &90.9 &97.9 &96.4 &99.0 &96.5 &\cellcolor{gray!35}95.8 \\
        FatFormer~\cite{liu2024forgery}  &\cellcolor{gray!35}93.3 &55.8 &92.7 &99.9 &100.0 &99.9 &75.9 &98.9 &98.8 &\cellcolor{gray!35}88.9 \\
        \midrule
        RAID (ours) &\cellcolor{gray!35}\underline{93.5} &\textbf{98.9}  &\textbf{97.2}  &97.7  &98.9  &98.8  &\textbf{97.8}  &\textbf{99.1}  &\textbf{97.5}  &\cellcolor{gray!35}\textbf{98.4} \\
        \bottomrule
    \end{tabular}}
    \caption{\textbf{Evaluation on AIGCDB~\cite{zhong2023patchcraft} and GenImage~\cite{zhu2024genimage}.} Following existing protocols, for AIGCDB, models are trained on ProGAN and evaluated on all subsets of AIGCDB, with the averaged accuracy reported; for GenImage, models are trained on Stable Diffusion V1.4 and evaluated on all subsets of GenImage.}
    \label{tab:comparison}
\end{table*}

\section{Experiments}
\label{sec:exp}
\subsection{Datasets and Implementation Details}
We conduct extensive experiments on various mainstream datasets, including AIGCDetectionBenchmark (AIGCDB)~\cite{zhong2023patchcraft}, GenImage~\cite{zhu2024genimage}, and our proposed GID and GVD. We adopt accuracy (ACC) as the evaluation metric.

\noindent\textbf{AIGCDB:} AI-generated images in the AIGCDB are generated by 17 GAN-based or Diffusion-based generators. We train the model on the subset ProGAN, and test on all 17 subsets to compute the average accuracy.

\begin{figure*}[htbp]
  \centering
  \small
  \includegraphics[width=0.95\linewidth]{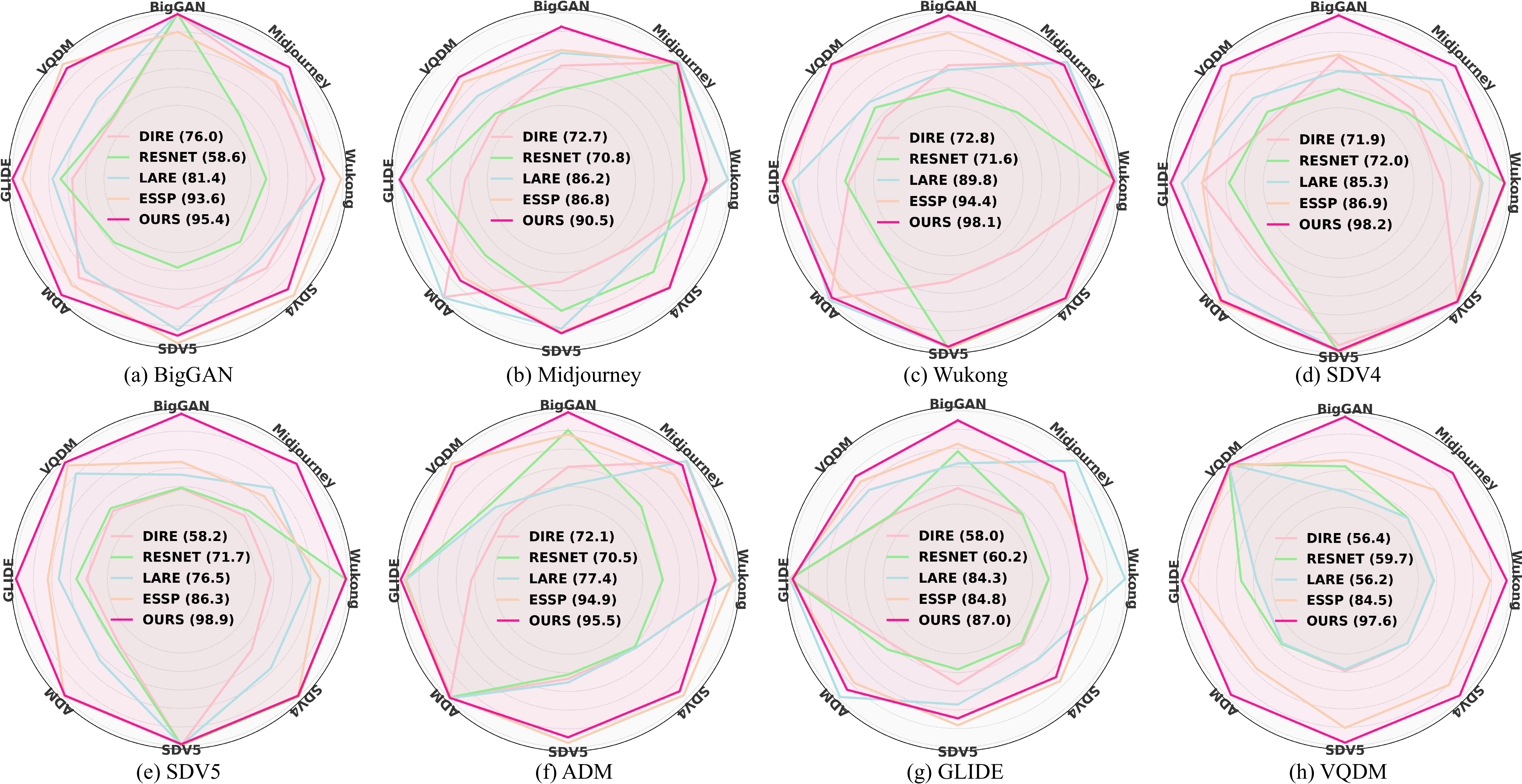}
  \caption{\textbf{Evaluation of generalization capability.} Four competitive methods and our approach are trained on eight subsets (corresponding to eight subplots) and evaluated on eight subsets (corresponding to eight dimensions of each subplot) of GenImage~\cite{zhu2024genimage}.}
  \label{fig:generalization}
\end{figure*}

\noindent\textbf{GenImage:} GenImage benchmark is a million-scale dataset especially designed for AI-generated image detection. Real images are sourced from ImageNet dataset~\cite{deng2009imagenet}, while AI-generated images are generated by eight mainstream GAN and Diffusion based generators, including BigGAN (BigG)\cite{BigGAN}, Midjourney (Midj)\cite{Midjourney}, Wukong (Wuk)\cite{wukong}, Stable Diffusion V1.4 (SDV4)\cite{rombach2022high}, Stable Diffusion V1.5 (SDV5)\cite{rombach2022high}, ADM \cite{dhariwal2021diffusion}, GLIDE (GLI)\cite{nichol2021glide}, and VQDM \cite{gu2022vectorvqdm}. 
Following the setting of~\cite{luo2024lare,yan2024sanity,chen2024single}, we train the model on subset Stable Diffusion V1.4, and test on all eight subsets.

\begin{table}[tbp]
    \centering
    \adjustbox{max width=0.45\textwidth}{
    \begin{tabular}{l|ccccccccc}
        \toprule
        Trainning &Big &Mid &Wuk &SD4 &SD5 &ADM &GLI &VQD &Avg. \\
        \midrule
        Big &99.6 &94.9 &86.0 &92.5 &92.7 &99.3 &99.7 &93.4 &\cellcolor{gray!35}94.7 \\
        Mid &84.5 &96.3 &79.2 &84.1 &84.9 &79.2 &93.2 &99.3 &\cellcolor{gray!35}85.0 \\
        Wuk &98.8 &95.3 &98.1 &98.4 &98.4 &97.7 &98.4 &98.1 &\cellcolor{gray!35}97.9 \\
        SD4 &98.4 &96.8 &97.5 &98.4 &98.4 &97.7 &98.6 &97.5 &\cellcolor{gray!35}98.0 \\
        SD5 &97.8 &95.9 &96.1 &97.6 &97.8 &96.9 &98.4 &96.2 &\cellcolor{gray!35}97.1 \\
        ADM &99.7 &95.6 &88.1 &93.8 &93.8 &99.8 &99.8 &94.7 &\cellcolor{gray!35}95.6 \\
        GLI &83.9 &82.8 &66.6 &71.4 &71.2 &81.9 &98.7 &95.2 &\cellcolor{gray!35}78.7 \\
        VQD &99.3 &94.8 &98.9 &99.1 &99.0 &98.4 &99.1 &99.6 &\cellcolor{gray!35}98.5 \\
        \midrule
        Average &95.3 &94.1 &88.8 &91.9 &92.0 &93.9 &98.2 &96.8 &\cellcolor{gray!35}93.2 \\
        \bottomrule
    \end{tabular}
    }
    \caption{\textbf{Cross-generator performance on the GenImage dataset.} We train our model on eight subsets of GenImage subsets respectively, and each model is evaluated on these eight subsets. The detection accuracy of both training subsets and testing subsets is averaged. For conciseness, BigGAN, Midjourney, Wukong, Stable Diffusion V1.4, Stable Diffusion V1.5, ADM, GLIDE, and VQDM are denoted as Big, Mid, Wuk, SD4, SD5, ADM, GLI, and VQD, respectively.}
    \label{tab:experiment}
\end{table}

\begin{table*}[ht]
    \centering
    \begin{minipage}[t]{\columnwidth} 
        \centering
        \adjustbox{max width=\textwidth}{
        \begin{tabular}{lcccccc}
        \toprule
        Method & Imagen 2 & FLUX.1 & DALL-E 3 & SD3 & WANX 2.1 & Avg.\\
        \midrule
        LGrad~\cite{tan2023learning} & 75.7 &57.0 &77.2 &82.5 &76.1 &\cellcolor{gray!35}73.7 \\
        LNP~\cite{liu2022detecting} & 82.4 &54.9 &9.1 &52.1 &43.7 &\cellcolor{gray!35}48.4 \\
        CNNSpot~\cite{wang2020cnn} &50.5 &49.2 &59.1 &66.2 &51.0 &\cellcolor{gray!35}55.2 \\
        GramNet~\cite{liu2020global}  &64.4 &49.4 &56.8 &70.3 &64.4 &\cellcolor{gray!35}61.1 \\
        ResNet-50~\cite{he2016deepresnet}  &54.9 &51.8 &63.6 &67.5 &61.6 &\cellcolor{gray!35}60.0 \\
        UnivFD~\cite{ojha2023towards} &49.7 &69.2 &50.0 &58.7 &50.8 &\cellcolor{gray!35}55.7 \\
        DIRE \cite{wang2023dire} &51.4 &50.0 &51.0 &77.2 &46.9 &\cellcolor{gray!35}55.3 \\
        LaRE$^2$~\cite{luo2024lare} &57.9 &81.1 &59.1 &60.0 &50.5 &\cellcolor{gray!35}61.7 \\
        ESSP~\cite{chen2024single} &\underline{96.7} &84.3 &81.8 &\underline{95.2} &\underline{83.8} &\cellcolor{gray!35}\underline{88.4} \\
        AIDE~\cite{yan2024sanity} &87.3 &\underline{91.0} &\underline{95.1} &86.6 &70.3 &\cellcolor{gray!35}86.1 \\
        \midrule
        RAID (ours) & \textbf{98.8} &\textbf{99.4}  &\textbf{97.4} &\textbf{98.9} &\textbf{94.2} &\cellcolor{gray!35}\textbf{97.7} \\
        \bottomrule
    \end{tabular}}
    \caption{\textbf{Cross-dataset evaluation on the proposed GID dataset.} We conduct cross-dataset deepfake detection evaluation on GID.}
    \label{tab:gid}
    \end{minipage}
    \hfill
    \begin{minipage}[t]{\columnwidth} 
        \centering
        \adjustbox{max width=\textwidth}{
        \begin{tabular}{l|cccccc}
        \toprule
        \small Group 1 &MuseV &SVD &Mora &CogV &HD-VG &Avg. \\
        \midrule
        \small LaRE$^2$~\cite{luo2024lare} &7.1 &6.8 &23.6 &37.5 &63.8 &\cellcolor{gray!35}41.3 \\
        \small ESSP~\cite{chen2024single}  &\underline{33.4} &\underline{38.4} &\underline{32.6} &\underline{39.8} &61.5 &\cellcolor{gray!35}\underline{48.8} \\
        \small AIDE~\cite{yan2024sanity} &15.7 &17.3 &27.2 &23.2 &\underline{75.2} &\cellcolor{gray!35}48.0 \\
         RAID (ours) &\textbf{44.5} &\textbf{53.6} &\textbf{62.4} &\textbf{62.4} &\textbf{77.2} &\cellcolor{gray!35}\textbf{66.5} \\
        \toprule
        \small Group 2 &COG  &T2VZ &TAV &VC &YT-BI &Avg. \\
        \midrule
        \small LaRE$^2$~\cite{luo2024lare} &13.1 &15.8 &32.1 &\textbf{38.0} &55.8 &\cellcolor{gray!35}40.3 \\
        \small ESSP~\cite{chen2024single} &\underline{28.8} &18.8 &15.4 &19.7 &47.4 &\cellcolor{gray!35}34.0 \\
        \small AIDE~\cite{yan2024sanity} &2.4 &\textbf{40.2} &\underline{37.4} &23.0 &\underline{76.3} &\cellcolor{gray!35}\underline{51.0} \\
         RAID (ours) &\textbf{35.1} &\underline{20.2} &\textbf{64.6} &\underline{34.8} &\textbf{79.5} &\cellcolor{gray!35}\textbf{59.1} \\
        \bottomrule
    \end{tabular}}
    \caption{\textbf{Cross-dataset evaluation on the proposed GVD dataset.} We conduct cross-dataset deepfake detection evaluation on GVD.}
    \label{tab:gvd}
    \end{minipage}
\end{table*}

\begin{table*}[ht]
    \centering
    \begin{minipage}[t]{\columnwidth} 
        \centering
        \adjustbox{max width=\textwidth}{
        \begin{tabular}{l|ccccccccc}
        \toprule
        Method &BigG &Midj &Wuk &SDV4 &SDV5 &ADM &GLI &VQDM &Avg. \\
        \midrule
        RIGID~\cite{he2024rigid} &53.0 &94.1 &87.8 &87.0 &87.2 &51.4 &45.9 &52.2 &\cellcolor{gray!35}69.8 \\
        AEROBLADE~\cite{ricker2024aeroblade} &58.3 &40.2 &51.4 &52.6 &55.1 &50.7 &29.4 &52.8 &\cellcolor{gray!35}48.8 \\
        Manifold~\cite{brokman2025manifold} &77.6 &55.5 &65.4 &62.0 &63.0 &57.3 &88.3 &76.9 &\cellcolor{gray!35}68.2 \\
        \midrule
        RAID (ours) & 91.0 &85.9 & 86.2 & 86.3 &86.8 &85.5 &83.9 &84.5 & \cellcolor{gray!35}86.3 \\
        \bottomrule
    \end{tabular}}
    \caption{\textbf{Zero-shot AI-generated image detection on GenImage.} We only utilize real images of ImageNet to construct zero-shot scenarios.}
    \label{tab:zero}
    \end{minipage}
    \hfill
    \begin{minipage}[t]{\columnwidth} 
        \centering
        \adjustbox{max width=\textwidth}{
        \begin{tabular}{l|ccccccccc}
        \toprule
        Method &BigG &Midj &Wuk &SDV4 &SDV5 &ADM &GLI &VQDM &Avg. \\
        \midrule
        RAID (ours) &\textbf{98.9}  &\textbf{97.2}  &97.7  &98.9  &98.8  &\textbf{97.8}  &\textbf{99.1}  &\textbf{97.5}  &\cellcolor{gray!35}\textbf{98.4} \\
        w/o BRI  &\underline{74.1} &\underline{90.9} &99.6 &\underline{99.9} &\textbf{99.9} &\underline{58.1} &\underline{90.5} &\underline{84.1} &\cellcolor{gray!35}\underline{87.7}\\
        w/o GBPS &57.0 &58.2 &\textbf{99.9} &\textbf{100.0} &\textbf{99.9} &55.0 &58.7 &60.3 &\cellcolor{gray!35}74.8 \\
        w/o BRI-GBPS &52.0 &54.9 &\underline{98.2} &\underline{99.9} &\underline{99.7} &53.5 &61.9 &56.2 &\cellcolor{gray!35}72.1 \\
        \bottomrule
    \end{tabular}}
    \caption{\textbf{Ablation studies.} BRI and GBPS modules are removed sequentially.}
    \label{tab:ablation}
    \end{minipage}
\end{table*}

\begin{table*}[ht]
    \centering
    \begin{minipage}[t]{\columnwidth} 
        \centering
        \adjustbox{max width=\textwidth}{
        \begin{tabular}{lcccccccc}
        \toprule
        Method & Gau-0 & Gau-1 &Gau-2 & Gau-3 &JP-100 & JP-98 &JP-95 &JP-90 \\
        \midrule
        RAID (ours) &98.4 &84.8  &80.4 &77.5 &98.4 &82.3  &79.5 &75.6\\
        ESSP~\cite{chen2024single} &89.7 &80.9  &58.3 &53.4 &89.7 &80.3  &74.5 &66.0\\
        UniFD~\cite{ojha2023towards} &70.1 &64.7 &64.1 &61.9 &70.1 &65.5 &64.5 &62.9\\
        \bottomrule
    \end{tabular}}
    \caption{\textbf{Results under image perturbations.} Different Gaussian blur (Gau) and JPEG compression (JP) are applied.}
    \label{tab:robustness}
    \end{minipage}
    \hfill
    \begin{minipage}[t]{\columnwidth} 
        \centering
        \adjustbox{max width=\textwidth}{
        \begin{tabular}{lcccc}
        \toprule
        \multirow{3}{*}{Method} & \multicolumn{2}{c}{Time} & \multicolumn{2}{c}{Params} \\
        \cmidrule(lr){2-3} \cmidrule(lr){4-5}
        & {\makecell{Feature Extraction}} & Total & {\makecell{Feature Extraction}} & Total\\
        \midrule
        DIRE~\cite{wang2023dire}        &1.99 s   &2 s      &644.8 M  &688.3 M \\
        LaRE$^2$~\cite{luo2024lare}     &250 ms   &260 ms   &1066.2 M &1165.8 M \\
        ESSP~\cite{chen2024single}      &25.10 ms &31.99 ms &7.1 M    &30.7 M \\
        \midrule
        RAID (ours)                     &2.09 ms  &4.23 ms  &0       &23.5 M \\
        \bottomrule
    \end{tabular}}
    \caption{\textbf{Comparison of computation efficiency.} Operation time and parameters are compared with other mainstream methods.}
    \label{tab:computation}
    \end{minipage}
\end{table*}

\noindent\textbf{GID and GVD:} We construct a new dataset to further evaluate the performance. Real images are sourced from ImageNet~\cite{deng2009imagenet}, and the AI-generated images are generated by currently competitive generators: Google Imagen 2, FLUX.1, DALL-E 3, Stable Diffusion 3 (SD3), and WANX 2.1. We also construct a challenging benchmark by employing AI-generated images extracted from AI-generated videos, which are generated by state-of-the-art video generative models including MuseV~\cite{musev}, SVD~\cite{svd}, Mora~\cite{mora}, CogVideo (CogV)\cite{cogvideo}, Text2Video-Zero (T2VZ)\cite{t2vz}, Tune-A-Video (TAV)\cite{wu2023tune}, and VideoCrafter2 (VC)\cite{vc2}. Real images are from videos including HD-VG~\cite{hd_vg_130m}, Youtube, and Bilibili (YT-BI). More details are shown in Tab~\ref{tab:challenging}.

We primarily use hard cases of AI-generated detection~\cite{ni2025genvidbenchchallengingbenchmarkdetecting}, such as plants, vehicles, people, buildings, natures, etc, with similar sample distribution across these categories. 
These datasets cover challenging scenarios such as low-illumination scenes (16.6$\%$, including twilight, dawn, nighttime, low-light environments, etc.), fast-moving objects (22.4$\%$, including vehicles, human motions, animal movements, natural phenomena, etc), and extreme environment scenes (4.9$\%$, including high-risk challenges, aerial activities, etc.). Details about our proposed datasets are summarized in Table~\ref{tab:challenging}.

\noindent\textbf{Implementation Details:} The image is first resized to 256$\times$256 before construction of bit-reversed image. Patches are randomly sampled during training. After the 32$\times$32 patch is selected, it is fed into the subsequent modified ResNet-50 classifier, which is pretrained on ImageNet~\cite{deng2009imagenet}. Training is conducted with a maximum of 16 epochs, a batch size of 64, a learning rate of 0.0001, and the Adam optimizer. For zero-shot scenarios, we only utilize real images of ImageNet~\cite{deng2009imagenet}. Apart from bit-reversed construction and patch selection, these images are fed into modified ResNet-50 pretrained on bit-reversed images to obtain the average feature after the final global average pooling. Then the testing images are input into the same architecture to obtain testing features for evaluating distances from features of real images.

\subsection{Evaluation of AI-Generated Image Detection}

\noindent\textbf{Evaluation on AIGCDB and GenImage:} We compare our results with other mainstream methods on three datasets, and results are shown in Table~\ref{tab:comparison}. 

On the AIGCDB dataset, our approach surpasses all prevailing methods, achieving an average accuracy of 93$\%$. 

On the GenImage dataset, despite marginally inferior performance on a few subsets, our method exhibits extremely strong generalization capabilities on all subsets, and achieves an average accuracy of 98.4$\%$, surpassing SOTA approaches by over 8.7$\%$ margin. 

\noindent\textbf{Cross-Generator Performance on the GenImage:}
We train our model on eight subsets of GenImage~\cite{zhu2024genimage}, and evaluate each model on eight subsets, respectively. As shown in Table~\ref{tab:experiment}, all of our models achieve an average accuracy of exceeding 90$\%$, except for the model trained on GLIDE attains the averaged performance of 87.0$\%$, possibly due to different feature distribution after bit-reversal. Notably, models encounter a slight decline when evaluated on the Wukong, demonstrating the challenge inherent in this subset.

\noindent\textbf{Cross-Dataset Evaluation on the proposed GID and GVD Dataset:} We train the model on Stable Diffusion V1.4 from GenImage, and test on all subsets of the proposed GID and GVD dataset.

On the proposed GID dataset, we reproduce the results of recent AI-generated image detection methods. As shown in Table~\ref{tab:gid}, most methods face substantial difficulties, with their accuracy dropping to around 50$\%$, highlighting the formidable challenge posed by the GID dataset. Although recent approaches AIDE and ESSP exceed 80$\%$ accuracy, they still significantly underperform our method achieving 97.7$\%$.

On the proposed GVD dataset, we evaluate LaRE$^2$, AIDE, ESSP and our method, all achieving high detecting accuracy on GenImage, on our challenging benchmark GVD. As shown in Table~\ref{tab:gvd}, all methods encounter a significant drop on GVD, which results from the fundamental distinction between the distribution of artifacts in fake images and videos, substantially validating the formidable challenge posed by our proposed dataset GVD for AI-generated image detection. Nonetheless, our approach evidently surpasses prevailing methods.

\noindent\textbf{Capability of Generalization Across Generators:} We compare our approach with four competitive methods (DIRE, ResNet-50, LaRE$^2$, and ESSP) on eight subsets of GenImage~\cite{zhu2024genimage}. As shown in Fig.~\ref{fig:generalization}, existing mainstream methods achieve reasonable results only when the training subset and testing subset are identical, and easily encounter difficulties when evaluated on subsets from unseen generators. Despite recent methods (e.g., LaRE$^2$ and ESSP) mitigating these generalization issues by achieving strong performance on subsets such as Wukong and Stable Diffusion V1.4 (V1.5), they still struggle to detect AI-generated images on other subsets, such as BigGAN, ADM, and Midjourney. Remarkably, our method, merely trained on one subset, shows exceptional detecting accuracy on all subsets, exhibiting excellent generalization capability.

\noindent\textbf{Zero-Shot Generalization Performance:} To further verify the effectiveness of our bit-reversed images, we conduct zero-shot AI-generated image detection using only real images from ImageNet for training. As there is limited prior work on this zero-shot setting, we compare our approach with three methods: RIGID~\cite{he2024rigid}, AEROBLADE~\cite{ricker2024aeroblade}, and Manifold~\cite{brokman2025manifold}. 
Results on the eight subsets of GenImage~\cite{zhu2024genimage} are shown in Table~\ref{tab:zero}. Without using AI-generated images for training, our RAID achieves 86.3\% average accuracy and strong performance across all eight subsets, whereas other mainstream zero-shot methods perform well on only a few subsets. 

\subsection{Ablation Studies and Analyses} 

We conduct extensive ablation studies and analyses to validate the effectiveness of our approach. All models are trained on Stable Diffusion V1.4, and evaluated on eight subsets of GenImage.

\noindent\textbf{Ablation Studies:} We respectively remove modules of Bit-Reversed Images (BRI), and Gradient-Based Patch Selection (GBPS) and compare results in Table~\ref{tab:ablation}. After removing the BRI and GBPS modules, the average accuracy drops from 98.4$\%$ to 87.7$\%$ and 74.8$\%$, respectively, highlighting the importance of each module in enhancing detection performance and generalization capability.

\begin{table*}[tbp]
    \centering
    \resizebox{\textwidth}{!}{
    \renewcommand{\arraystretch}{0.8}
    \scriptsize
    \begin{tabular}{l l c c c c c c c c c c c c c c c c c}
        \toprule
        \multicolumn{2}{c}{Method} &\multicolumn{8}{c}{Bit Order: From 0 to 7} &Big &Mid &Wuk &SD4 &SD5 &ADM &GLI &VQD &Avg. \\
        \midrule
        \multirow{8}{*}{\rotatebox{90}{Bit-Reversed}} &\textcircled{1} &7 &6 &5 &4 &3 &2 &1 &0 &\textbf{98.9}  &97.2  &\textbf{97.7}  &\textbf{98.9}  &\textbf{98.8}  &\textbf{97.8}  &\textbf{99.1}  &\textbf{97.5}  &\cellcolor{gray!35}\textbf{98.4} \\
        &\textcircled{2} &6 &5 &4 &3 &2 &1 &0 &7 &92.5 &92.2 &94.2 &97.3 &97.4 &87.7 &97.8 &88.3 &\cellcolor{gray!35}93.6 \\
        &\textcircled{3} &5 &4 &3 &2 &1 &0 &7 &6 &90.9 &81.9 &92.5 &95.2 &95.2 &74.2 &75.3 &91.7 &\cellcolor{gray!35}87.4 \\
        &\textcircled{4} &4 &3 &2 &1 &0 &7 &6 &5 &90.7 &92.2 &89.2 &94.3 &94.2 &76.0 &82.0 &89.3 &\cellcolor{gray!35}88.7 \\
        &\textcircled{5} &3 &2 &1 &0 &7 &6 &5 &4 &\underline{97.7} &\textbf{98.9} &94.5 &\underline{97.9} &\underline{98.2} &88.2 &89.7 &94.3 &\cellcolor{gray!35}95.1 \\
        &\textcircled{6} &2 &1 &0 &7 &6 &5 &4 &3 &97.4 &97.1 &\underline{95.9} &97.2 &97.1 &\underline{97.2} &97.7 &\underline{96.5} &\cellcolor{gray!35}\underline{97.0} \\
        &\textcircled{7} &1 &0 &7 &6 &5 &4 &3 &2 &96.9 &\underline{97.5} &92.7 &\underline{97.9} &98.0 &88.2 &\underline{99.0} &90.7 &\cellcolor{gray!35}95.2 \\
        &\textcircled{8} &0 &7 &6 &5 &4 &3 &2 &1 &89.8 &95.5 &92.4 &96.4 &96.3 &82.0 &88.3 &89.0 &\cellcolor{gray!35}91.4 \\
        \midrule
        \multirow{8}{*}{\rotatebox{90}{Bit-Forward}} & \textbf{\textcircled{1}} & 0 &1 &2 &3 &4 &5 &6 &7 &74.1 &90.9 &\textbf{99.6} &\textbf{99.9} &\textbf{99.9} &58.1 &90.5 &84.1 &\cellcolor{gray!35}87.7\\
        &\textbf{\textcircled{2}} &1 &2 &3 &4 &5 &6 &7 &0 &81.4 &86.5 &95.1 &98.3 &98.6 &60.1 &81.7 &82.4 &\cellcolor{gray!35}86.0\\
         &\textbf{\textcircled{3}} &2 &3 &4 &5 &6 &7 &0 &1 &91.1 &91.9 &91.9 &96.6 &96.6 &63.5 &80.6 &86.2 &\cellcolor{gray!35}87.7\\
        &\textbf{\textcircled{4}} &3 &4 &5 &6 &7 &0 &1 &2 &\underline{97.3} &95.2 &96.5 &98.8 &98.7 &86.6 &82.3 &97.1 &\cellcolor{gray!35}94.3\\
        &\textbf{\textcircled{5}} &4 &5 &6 &7 &0 &1 &2 &3 &96.6 &\textbf{98.9} &94.0 &98.1 &98.3 &83.4 &83.4 &93.3 &\cellcolor{gray!35}93.5\\
        &\textbf{\textcircled{6}} &5 &6 &7 &0 &1 &2 &3 &4 &\textbf{98.5} &\underline{98.3} &97.1 &98.3 &98.4 &\textbf{97.3} &\textbf{98.7} &\underline{98.7} &\cellcolor{gray!35}\textbf{98.0}\\
        &\textbf{\textcircled{7}} &6 &7 &0 &1 &2 &3 &4 &5 &95.2 &97.7 &93.6 &98.0 &97.8 &\underline{87.2} &\textbf{98.7} &\textbf{98.8} &\cellcolor{gray!35}\underline{95.0}\\
        &\textbf{\textcircled{8}} &7 &0 &1 &2 &3 &4 &5 &6 &90.5 &93.3 &\underline{98.9} &\underline{99.7} &\underline{99.7} &77.1 &\underline{98.3} &82.4 &\cellcolor{gray!35}92.8\\
        \bottomrule
    \end{tabular}}
    \caption{\textbf{Performances of different bit forward and reversed images.} Different bit forward and reversed images are utilized. Our approach is \textcircled{1} of bit-reversed images, and \textbf{\textcircled{1}} of bit-forward images is the baseline using the original RGB image.}
    \label{tab:reverse}
\end{table*}

\noindent\textbf{Performances of Different Bit-Forward and Bit-Reversed Images:} To determine which bit-forward or bit-reversed image is the most effective, we compare results of diverse bit-forward and bit-reversed images in Table~\ref{tab:reverse}.
In terms of bit-reversed images, \textcircled{1} represents full reversal and exhibits the most competitive results of 98.4$\%$, demonstrating the effectiveness of prioritizing lower-order bit-planes.
In terms of bit-forward images, average accuracies achieve a high level for variants \textbf{\textcircled{4}}$\sim$\textbf{\textcircled{7}}, when more lower-order bit-planes are moved to higher-order positions. Notably, it reaches the peak of 98.0$\%$ for the \textbf{\textcircled{6}}.

\noindent\textbf{Impact of Different Weights}
Table~\ref{tab:reweighting} further investigates the impacts of different weights while constructing bit reversed images. Generally, we assign most significant weights to the lowest-order bit-planes (such as 0$\sim$1, 0$\sim$2, 1$\sim$3, etc) based on results of Table 6. The variant \textcircled{1} denotes our approach, while \textcircled{8} is the variant with learnable weights during training. Comparing \textcircled{1} with \textcircled{2}$\sim$\textcircled{4}, we observe that increasing the weights of higher bit-planes leads to a noticeable degradation in performance. In contrast, while comparing \textcircled{1} with \textcircled{5}, we find that increasing the weights of lower bit-planes slightly increases the performance. These experiments further highlight the critical role of low-bit planes for AI-generated image detection. Although the variant of learnable weights eliminates the need for handcrafted weights and manual tuning, it performs worse than the others, possibly due to overfitting and the model not recognizing the importance of lower bit-planes. Although \textcircled{5} performs even better than ours (\textcircled{1}), we use the default weights in our approach for simplicity.

\begin{table*}[tbp]
    \centering
    \resizebox{\textwidth}{!}{
    \begin{tabular}{l c c c c c c c c c c c c c c c c c}
        \toprule
     & 0 & 1 & 2 & 3 & 4 & 5 & 6 & 7 & Big & Mid & Wuk & SD4 & SD5 & ADM & GLI & VQD & Avg. \\
        \midrule
        \textcircled{1} &128 &64 &32 &16 &8 &4 &2 &1 &\underline{98.9} &97.2  &97.7  &\underline{98.9}  &98.8  &97.8  &\underline{99.1}  &97.5  &\cellcolor{gray!35}98.4 \\
        \textcircled{2} &48 &48 &48 &48 &32 &16 &8 &4 &90.4 &93.5 &91.5 &94.4 &93.4 &93.0 &94.6 &97.7 &\cellcolor{gray!35}91.3 \\
        \textcircled{3} &64 &64 &64 &32 &16 &8 &4 &2 &94.1 &95.9 &94.9 &96.3 &96.2 &87.5 &95.8 &92.2 &\cellcolor{gray!35}94.2 \\
       \textcircled{4} &32 &64 &64 &64 &16 &8 &4 &2 &92.9 &96.9 &93.0 &95.3 &95.4 &85.2 &96.5 &90.6 &\cellcolor{gray!35}93.3 \\
       \textcircled{5} &96 &96 &32 &16 &8 &4 &2 &1 &\textbf{99.4} &\textbf{98.5} &\textbf{99.0} &\textbf{99.6} &\textbf{99.5} &\textbf{99.3} &\textbf{99.6} &\textbf{99.0} &\cellcolor{gray!35}\textbf{99.3} \\
        \textcircled{6} &32 &96 &96 &16 &8 &4 &2 &1 &98.8 &\underline{98.3} &\underline{98.2} &\underline{98.9} &\underline{99.0} &\underline{98.5} &\underline{99.1} &\underline{98.0} &\cellcolor{gray!35}\underline{98.6} \\
        \textcircled{7} &32 &16 &96 &96 &8 &4 &2 &1 &94.6 &97.6 &93.5 &96.5 &96.6 &86.0 &96.3 &91.7 &\cellcolor{gray!35}94.2 \\
         \midrule
        \textcircled{8} &\multicolumn{8}{c}{Learnable} &85.6 &70.1 &94.7 &95.4 &95.1 &66.5 &91.2 &57.0 &\cellcolor{gray!35}82.2 \\
        \bottomrule
    \end{tabular}}
    \caption{\textbf{Impact of different weights of bit-planes.} Different and learnable weights are utilized during construction of bit reversed images. The column index denotes the position of bit-planes, and circled indices are different variants of our approach.}
    \label{tab:reweighting}
\end{table*}

\begin{figure*}[t]
  \centering
  \small
  \vspace{-1mm}
  \includegraphics[width=0.9\linewidth]{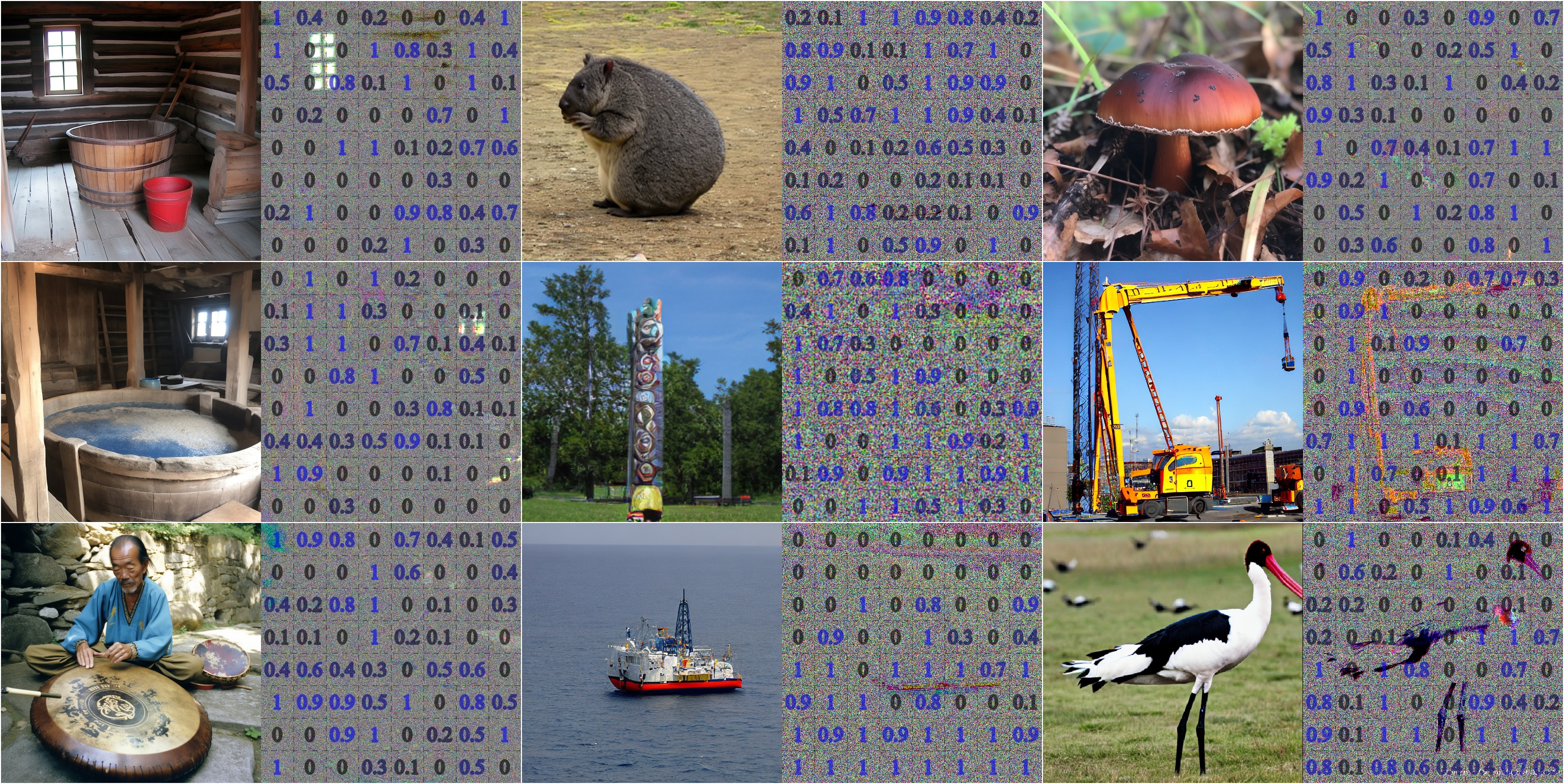}
  \vspace{-1mm}
  \caption{\textbf{Visualizations of AI-generated images and the corresponding probabilities of small patches predicted as fake}. The images are resized to 256×256, with a patch size of 32×32. The predicted fake probabilities are at the center of each patch on the right of RGB images, with more intense blue colors representing patches that are successfully predicted as fake.}
  \label{fig:final_vis}
\vspace{-3mm}
\end{figure*}

\noindent\textbf{Robustness Against Image Degradation:} In Table~\ref{tab:robustness}, we show detection results of RAID against two representative types of image degradation: Gaussian blur and JPEG compression. Compared to the single patch-based ESSP~\cite{chen2024single} and frequency-based UniFD~\cite{ojha2023towards}, our patch-based method using the bit-reversed image is more robust to image degradation and noise perturbation.

\noindent\textbf{Computation Efficiency:} We compare the computational efficiency of our approach with other mainstream methods (DIRE, LaRE$^2$, and ESSP) in Table~\ref{tab:computation}. For the inference speed, DIRE and LaRE require multiple steps to construct the feature map, resulting in higher latency (1.99 s and 250 ms, respectively), while our approach completes this process in a single step, taking only 2.09 ms. The patch-based method ESSP takes a total of 31.99 ms, while our approach operates at the millisecond level.
In terms of model parameters, other mainstream methods mainly rely on large pretrained models (e.g., diffusion), which introduce a substantial number of parameters. In contrast, our approach is significantly more lightweight and efficient, requiring only 23.5 M parameters.

\noindent\textbf{Visualizations of Fake Probabilities of Patches:} Since our approach is based on single patch of bit-reversed image, we visualize the predicted fake probabilities for evenly divided patches in Fig.~\ref{fig:final_vis}. We select AI-generated images that appear highly realistic, and visualize the predicted fake probabilities of patches of bit revised images. We find that artifacts in AI-generated images are invisible in the original images, but become visible in bit-reversed images and can be successfully detected by our model.

\section{Conclusion}
In this paper, we studies AI-generated image detection from the perspective of bit-planes and introduce an innovative representation named bit-reversed image. The bit-reversed image is a reversible encoding of the original RGB image, but it evidently amplifies artifacts that are invisible in the original image. Following this insight, we propose a simple yet highly effective approach for AI-generated image detection. 
Extensive experiments, including cross-generator evaluation, cross-dataset evaluation, and zero-shot AI-generated image detection, consistently demonstrate the effectiveness of our approach.
In addition, it contains only 23.5 million parameters and runs in milliseconds. One limitation of our approach is that the deepfake image classification model we use is the standard ResNet. We will design a more tailored architecture in the future.

{
    \small
    \bibliographystyle{ieeenat}
    \bibliography{main}
}


\end{document}